\pdfoutput=1
\documentclass[11pt]{article}

\usepackage{emnlp2023}

\usepackage{times}
\usepackage{latexsym}

\usepackage[T1]{fontenc}
\usepackage[utf8]{inputenc}

\usepackage{microtype}

\usepackage{inconsolata}

\usepackage{xurl} % Makes URLs in bibliography break nicely
\usepackage{multirow}
\usepackage{tabularx}
\usepackage{tabulary}
\usepackage{enumitem}
\usepackage{amsfonts}
\usepackage{amsmath}
\usepackage[nameinlink,capitalise]{cleveref}
\usepackage{graphicx}
\usepackage{calc}
\usepackage{changepage}
\usepackage{booktabs}
\usepackage{wrapfig}
\usepackage{caption}
\usepackage{subcaption}
\usepackage{ragged2e}
\usepackage{lipsum}
\usepackage[normalem]{ulem}
\usepackage{overpic}
\usepackage[many]{tcolorbox}
\usepackage{listings}
\usepackage{siunitx}
\usepackage{adjustbox}
\usepackage{xspace}
\usepackage{pdflscape}
\usepackage{tikz}
\usepackage{svg}
\usepackage{mdframed}
\usepackage{wasysym}
\usepackage{forloop}
\usepackage{xparse}
\usepackage{colortbl}
\usepackage{fancyvrb}
\usepackage{relsize}
\usepackage{soul}

\Crefname{table}{Table}{Tables}
\Crefname{figure}{Figure}{Figures}
\Crefname{equation}{Equation}{Equations}
\Crefname{part}{Part}{Parts}
\Crefname{section}{Section}{Sections}
\Crefname{subsection}{Section}{Sections}
\Crefname{subsubsection}{Section}{Sections}
\Crefname{lstlisting}{Algorithm}{Algorithms}

\crefname{table}{Table}{Tables}
\crefname{figure}{Figure}{Figures}
\crefname{equation}{Equation}{Equations}
\crefname{part}{Part}{Parts}
\crefname{section}{Section}{Sections}
\crefname{subsection}{Section}{Section}
\crefname{subsubsection}{Section}{Section}
\crefname{lstlisting}{Algorithm}{Algorithms}

\newcommand{\hquad}{\;\:}

\sethlcolor{Snow2}
\newcommand{\prompt}[1]{{\small\texttt{\hl{#1}}}}

\def\arraystretch{1.1}        % for the vertical padding
\setlist[itemize]    {leftmargin=0.5cm,itemsep=0pt,topsep=0pt,parsep=0pt,partopsep=0pt}
\setlist[enumerate]  {leftmargin=0.5cm,itemsep=0pt,topsep=0pt,parsep=0pt,partopsep=0pt}
\setlist[description]{leftmargin=0pt,itemsep=0pt,topsep=0pt,parsep=0pt,partopsep=0pt}

\makeatletter
\newcommand{\cleantitle}[1]{\def\@cleantitle{#1}}
\AtBeginDocument{\hypersetup{pdftitle=\@cleantitle}}
\newcommand{\cleanauthor}[1]{\def\@cleanauthor{#1}}
\AtBeginDocument{\hypersetup{pdfauthor=\@cleanauthor}}
\makeatother

\newcommand{\custombox}[3]{%
    \begingroup
    \colorlet{headercolor}{#1!80}
    \colorlet{bordercolor}{#1!60}
    \colorlet{rowcolor1}{#1!20}
    \colorlet{rowcolor2}{#1!40}
    \RaggedRight% align cell contents left
    \noindent%
    \begin{tcolorbox}[enhanced, colback=white, sharp corners, frame hidden, boxrule=0pt,
        borderline north={2pt}{0pt}{bordercolor},
        borderline south={2pt}{0pt}{bordercolor},
        left=0mm, right=0mm, top=2pt, bottom=2pt, boxsep=0mm]
        \begin{minipage}{\linewidth}
            \rowcolors{1}{rowcolor1}{rowcolor2} % alternating row colors
            \begin{tabularx}{\linewidth}{lX}
                #3
            \end{tabularx}
        \end{minipage}
    \end{tcolorbox}
    \endgroup
}

\newcommand{\contentrow}[2]{%
        {\small\textbf{#1}} & {\small#2} \\
}
\definecolor{MiroBlack}{HTML}{1a1a1a}
\definecolor{PredictionPipelineColor}{HTML}{7AB635}
\definecolor{KwEmbeddingPipelineColor}{HTML}{3C9CDB}
\definecolor{BabelnetEmbeddingPipelineColor}{HTML}{F24726}
\definecolor{OntoReplacePipelineColor}{HTML}{0CA789}
\colorlet{SummaryBoxColor}{Snow3}

\makeatletter
\newtcbox{\kwColorBox}[1][]{on line,fontupper=\footnotesize\sffamily\bfseries\small,boxrule=0.5pt,arc=2pt,coltext=#1,colback=#1!10!white,colframe=#1,boxsep=0pt,left=1.5pt,right=1.5pt,top=1.5pt,bottom=1.5pt}
\makeatother

\newcommand{\kw}[2]{%
    \begin{kwColorBox}[#2]%
    {#1}%
    \end{kwColorBox}%
    \xspace%
}

\renewcommand{\paragraph}[1]{\refstepcounter{paragraph}\noindent\textbf{#1\ ---}\label{par:\theparagraph}}

\definecolor{ColorUser}{HTML}{A42C2C}
\definecolor{ColorChallenge}{HTML}{A02FA5}
\definecolor{ColorLmChallenge}{HTML}{6B4E90}
\definecolor{ColorTask}{HTML}{408E2F}
\definecolor{ColorLmTask}{HTML}{AA7A39}
\definecolor{ColorWidget}{HTML}{4F85C3}
\newcommand{\rawuserdonotuse}[1]{\kw{#1}{ColorUser}}

\newcommand{\refuser}[1]{\hyperref[user:#1]{\rawuserdonotuse{#1}}}

\newcommand{\rawchallengedonotuse}[1]{\kw{#1}{ColorChallenge}}

\newcommand{\refchallenge}[1]{\hyperref[challenge:#1]{\rawchallengedonotuse{#1}}}

\newcommand{\rawlmchallengedonotuse}[1]{\kw{#1}{ColorLmChallenge}}
\newcommand{\deflmchallenge}[1]{\rawlmchallengedonotuse{\phantomsection\label{lmchallenge:#1}#1}}
\newcommand{\reflmchallenge}[1]{\hyperref[lmchallenge:#1]{\rawlmchallengedonotuse{#1}}}

\newcommand{\rawtaskdonotuse}[1]{\kw{T#1}{ColorTask}}

\newcommand{\reftask}[1]{\hyperref[task:#1]{\rawtaskdonotuse{#1}}}

\newcommand{\rawlmtaskdonotuse}[1]{\kw{#1}{ColorLmTask}}
\newcommand{\deflmtask}[1]{\rawlmtaskdonotuse{\phantomsection\label{lmtask:#1}#1}}
\newcommand{\reflmtask}[1]{\hyperref[lmtask:#1]{\rawlmtaskdonotuse{#1}}}

\newcommand{\rawwidgetdonotuse}[1]{\kw{W#1}{ColorWidget}}

\newcommand{\refwidget}[2]{\hyperref[widget:#1]{\rawwidgetdonotuse{#2}}}
\definecolor{CustomLstBackground}{HTML}{EAEAEA}
\definecolor{CustomLstForeground}{HTML}{212121}
\definecolor{CustomLstGreen}{HTML}{10a778}
\definecolor{CustomLstPurple}{HTML}{523c79}
\definecolor{CustomLstYellow}{HTML}{124D96}
\definecolor{CustomLstBlue}{HTML}{008ec4}
\lstdefinestyle{PythonCustomLst}{
    belowcaptionskip=1\baselineskip,
    breaklines=false,
    language=Python,
    showstringspaces=false,
    basicstyle=\tiny\ttfamily\color{CustomLstForeground},
    keywordstyle=\bfseries\color{CustomLstGreen},
    commentstyle=\itshape\color{CustomLstPurple},
    identifierstyle=\color{CustomLstYellow},
    stringstyle=\color{CustomLstBlue},
    backgroundcolor = \color{CustomLstBackground}
}

\newlength\myheight%
\newlength\mydepth%
\settototalheight\myheight{Xygp}%
\title{\textit{Revealing the Unwritten}: Visual Investigation of Beam Search Trees\\to Address Language Model Prompting Challenges}
\cleantitle{Revealing the Unwritten: Visual Investigation of Beam Search Trees to Address Language Model Prompting Challenges}

\author{
    \vphantom{O}\\[-0.8em]
    \textbf{
        Thilo Spinner\textsuperscript{1},\quad
        Rebecca Kehlbeck\textsuperscript{1},\quad
        Rita Sevastjanova\textsuperscript{1},\quad
        Tobias Stähle\textsuperscript{1},
    }\\
    \textbf{
        Daniel A. Keim\textsuperscript{1},\quad
        Oliver Deussen\textsuperscript{1},\quad
        Andreas Spitz\textsuperscript{1},\quad
        Mennatallah El-Assady\textsuperscript{2}
    }\\[0.2em]
    \textsuperscript{1}University of Konstanz\quad
    \textsuperscript{2}ETH Zürich\\
    \small\texttt{\{thilo.spinner, rebecca.kehlbeck, rita.sevastjanova, tobias.staehle, keim,}\\[-0.2em]
    \small\texttt{oliver.deussen, andreas.spitz\}@uni-konstanz.de,\quad{}menna.elassady@ai.ethz.ch}
}
\cleanauthor{Thilo Spinner, Rebecca Kehlbeck, Rita Sevastjanova, Tobias Stähle, Daniel A. Keim, Oliver Deussen, Andreas Spitz, and Mennatallah El-Assady }

\begin{document}

\maketitle
\begin{abstract}
    The growing popularity of generative language models has amplified interest in interactive methods to guide model outputs.
    Prompt refinement is considered one of the most effective means to influence output among these methods.
    We identify several challenges associated with prompting large language models, categorized into data- and model-specific, linguistic, and socio-linguistic challenges.
    A comprehensive examination of model outputs, including runner-up candidates and their corresponding probabilities, is needed to address these issues.
    The beam search tree, the prevalent algorithm to sample model outputs, can inherently supply this information.
    Consequently, we introduce an interactive visual method for investigating the beam search tree, facilitating analysis of the decisions made by the model during generation.
    We quantitatively show the value of exposing the beam search tree and present five detailed analysis scenarios addressing the identified challenges.
    Our methodology validates existing results and offers additional insights.
\end{abstract}

\section{Introduction}
Large language models (LLMs) have emerged as indispensable tools for text generation, and their aptitude for generating human-like text~\cite{Li2021PretrainedLanguageModel}, ease of use, and the wide range of application scenarios have pushed generative models into the general public.
The main lever to refine and steer the outputs of these models is the prompt, i.e., the model's initial input based on which new tokens are generated.
Many applications, therefore, focus on prompt engineering to steer results in the direction desired by the user~\cite{Webson2022DoPromptBased}.
However, comprehending the created outputs remains challenging for natural language processing (NLP) practitioners and linguistic experts.
Previous work has sought to address these challenges, with some efforts focusing on the explainability of LLMs \cite{Strobelt2018Seq2seqVisVisual,Lee2017InteractiveVisualizationManipulation,Strobelt2022GenniHumanAi}.
Complex behaviors and unwanted artifacts, such as biases and prompt sensitivity, typically hidden within the black-box nature of these models, have substantial implications for their usability and interpretability \cite{Alba2022OpenaiChatbotSpits,Ji2023SurveyHallucinationNatural}.
Most related works focus on explaining in which step problems occur and offer solutions to directly improve the created output for a specific task, such as machine translation.
However, they do not enable the user to deeply investigate phenomena in the entirety of the possible output space of the generative model.

To address this problem, we identify concrete \textit{prompting challenges}, covering data and model-specific, linguistic, and socio-linguistic aspects that may afflict the models' outputs.
The overarching tasks necessary to solve these challenges implicate that the user needs to explore probabilities of generated text, investigate alternative runner-up candidates, and allow for the comparison of different prompt variations -- all under the common theme of supporting explainability of the outputs.
Evaluating if (and how severely) a model is affected by a prompting challenge based solely on the generated output is not feasible using standard quantitative evaluation metrics since pruned candidates cannot be taken into consideration.
Therefore, we propose to analyze the output space of the model using the beam search tree representation to guide the user in identifying and tackling prompting challenges.

Used as part of the decision layer, the beam search tree (BST) generates possible hypotheses of outputs using the predicted token probabilities.
Analyzing its outputs per se poses a challenge since the tree may grow large and become cluttered, depending on the beam's width and the prediction's length.
To address this issue, we propose a visual approach that visually presents the beam search tree as the integral visualization workspace.
It allows NLP practitioners and linguistic experts to visually investigate the BST, enabling a direct comparison of prompt variations, semantic augmentations, and interactive adaptations of the output.

\noindent
Summarizing our contributions, we
\begin{itemize}
    \item identify and structure open challenges in the prompting of SOTA generative models;
    \item present a BST-based visual analytics technique and -workspace\footnote{The workspace will be made available upon acceptance.}, tailored to identify and address such challenges;
    \item quantitatively evaluate our tree-based approach;
    \item show how our tool can be applied to different scenarios tackling the identified challenges.
\end{itemize}

\section{Identifying Prompting Challenges}
\label{sec:problem-characterization-and-methodology}
Despite the recent success of large language models for text generation, several challenges remain elusive for data-driven solutions (in contrast to rule-based models).
In particular, we focus on challenges stemming from syntactic and semantic nuances in the input prompt as the user's main lever for influencing the output of a generative model.
In the following, we identify five prototypical, concrete challenges in utilizing deep learning-based, generative language models, which we derive from the state-of-the-art in literature, motivated by discussions with (computer) linguistic experts.
The identified challenges can be categorized into \textbf{data- and model-specific}, \textbf{linguistic}, and \textbf{socio-linguistic} challenges.

The challenges aim at NLP practitioners, who assess, employ, and fine-tune language models for NLP tasks, and linguistic experts, who investigate linguistic questions using language models.

\subsection{Data- \& Model-Specific Challenges}

Some characteristics of large language models are influenced by the pre-processing of training data and how the model is fine-tuned to a certain task (\emph{data-specific}).
Other challenges are inherent to the manner in which a model predicts its outputs and how these outputs are sampled during text generation (\emph{model-specific}).

\paragraph{Prompt Sensitivity~\deflmchallenge{Sens}}
The output of generative LMs is often sensible to small changes in the prompts, such as nuances in spacing or format (punctuation) or differences in the word order (syntax) in semantically similar sequences~\cite{Webson2022DoPromptBased}.
By semi-automatically varying the prompt and generating alternative trees for each variation, our approach can help in evaluating a model's sensitivity to prompts.

\paragraph{Surface Form Competition~\deflmchallenge{SFC}}
Distinctive to statistical models is the \emph{surface form competition}~\cite{Holtzman2021SurfaceFormCompetition}, in which the probability mass is distributed over multiple semantically equivalent words for the same underlying concept, consequently lowering the overall output probability of any correct token.
Our approach tackles surface form competition by communicating probabilities of alternative words to the user.

\subsection{Linguistic Challenges}

We define syntactic and semantic linguistic phenomena that are known to be hard to capture for LLMs as \emph{linguistic challenges}.

\paragraph{Negation~\deflmchallenge{Neg}}
Large language models are known to struggle with negation and negative imperatives, which has been shown for masked~\cite{Kassner2020NegatedMisprimedProbes,Kalouli2022NegationCoordinationQuantifiersa} and generative models~\cite{SummersStay2021WhatCanGenerative,Truong2023LanguageModelsAre}.
How these models capture negation is typically investigated by analyzing the model's \textit{top} prediction (see, e.g.,~\citet{SummersStay2021WhatCanGenerative}).
Using prediction \textit{alternatives} (i.e., top-$k$ predictions), we demonstrate that some models do not just ignore the inclusion of negative imperatives in the prompt but even boost the probabilities of undesired tokens.

\paragraph{Quantifiers~\deflmchallenge{Quant}}
How LLMs capture the semantics of quantifiers is of linguistic interest and has been investigated for masked language models~\cite{Warstadt2019InvestigatingBertsKnowledge,Kalouli2022NegationCoordinationQuantifiersa} and generative models.
In particular,~\citet{Gupta2023ProbingQuantifierComprehension} showed that larger generative models encode quantifiers better than smaller models.
Using BST exploration, we demonstrate how the output for near identical prompts with quantifier variations can be investigated effectively.

\subsection{Socio-Linguistic Challenges}

\paragraph{Bias~\deflmchallenge{Bias}}
Bias is a major challenge data-driven language models face, and numerous approaches for its detection and mitigation have been proposed \cite{Mehrabi2021SurveyBiasFairness}.
While there have been successes, methods have been criticized for inconsistent measurements \cite{Husse2022MindYourBias} and a lack of adherence to real-world biases~\cite{Blodgett2020LanguageTechnologyIs}.
Since the analysis of biases in text generation can be nuanced, and biases may arise during the generation of any token \cite{Liang2021TowardsUnderstandingMitigating},  the task is sensitive to the design of template prompts, meaning that template-based prompts may evoke biases itself \cite{Alnegheimish2022UsingNaturalSentence}.
To support the development of rigorous detection methods, we propose a tree-based approach for comparative, exploratory bias analysis, allowing the detection of biases in variable-length sequences and the identification of subtle nuances in the models' predictions.
We show how our tool can reveal model biases by comparing instance-based tree alternatives.

\section{The generAItor Workspace}
\label{sec:generaitor-workspace}
In this section, we briefly describe the generAItor workspace that we use for BST exploration of prompting challenges.
The workspace provides a visual interactive interface for loading language models, configuring beam search parameters, generating text, and investigating and comparing the generated beam search trees.

\subsection{User Tasks}
\label{subsec:tasks}

To tackle the identified prompting challenges, we consider the following tasks that the user has to perform.
They ground the design of generAItor, to enable the generation and investigation of BSTs based on different models and prompts.

\paragraph{Configuration \deflmtask{Conf}}
To compare different transformer-based LLMs, loading models and adjusting beam search parameters are required.

\paragraph{Text Generation \deflmtask{Gen}}
Users can specify a starting prompt.
Text is generated using the prompt, model, and beam search parameters.

\paragraph{Single-Instance Analysis \deflmtask{Single}}
To investigate a single BST instance, the user needs to explore alternative paths, assess output probabilities, and identify content similarity, undesired patterns, and sentiment changes.
As an example of a single-instance analysis, consider an investigation of the semantic constraint of the negation ``not.''
The user would define a prompt for an instruction model with ``do not use the following word \textit{x}'' and observe the probability of the undesired output in the BST.

\paragraph{Multi-Instance Analysis \deflmtask{Multi}}
To compare multiple BST instances, tree variations based on template prompts need to be generated automatically so that the user can observe syntactic and semantic differences in the trees.
E.g., using the negation example, the user could define a prompt including ``do not use the following word \textit{[x,y,z]}'' and compare the three resulting BST instances.

\begin{table*}[!htb]
    \setlength{\tabcolsep}{0.32em}   % adjust the horizontal padding
    \renewcommand*\arraystretch{1.1} % adjust the vertical padding
    \scriptsize
    \centering
    \begin{tabular}{@{}c|cccccccccccc|cccccccccccc@{}}
    \toprule
    \multirow{2}{*}{\textbf{Prompt}} & \multicolumn{12}{c|}{\multirow{2}{*}{\texttt{\textless{}John,Jessica\textgreater{} works as {[}Occupations{]}}}}                                                                                                                                                                                                                                                                                                  & \multicolumn{12}{c}{\multirow{2}{*}{\parbox{7cm}{\centering\texttt{World economy is strongly dependent of some countries, such as {[}Countries{]}}}}} \\
    & \multicolumn{12}{c|}{}                                                                                                                                                                                                                                                                                                                                                                                          & \multicolumn{12}{c}{}                                                                                                                                                                    \\
    \midrule
    \textbf{Model}                   & \multicolumn{6}{c|}{bloom-3b}                                                                                                              & \multicolumn{6}{c|}{RedPajama-INCITE-Base-3B-v1}                                                                       & \multicolumn{6}{c|}{bloom-3b}                                                                                                              & \multicolumn{6}{c}{RedPajama-INCITE-Base-3B-v1}                                                                       \\
    \midrule
    \textbf{n}                       & \multicolumn{2}{c|}{25}                      & \multicolumn{2}{c|}{50}                      & \multicolumn{2}{c|}{100}                     & \multicolumn{2}{c|}{25}                      & \multicolumn{2}{c|}{50}                      & \multicolumn{2}{c|}{100} & \multicolumn{2}{c|}{25}                      & \multicolumn{2}{c|}{50}                      & \multicolumn{2}{c|}{100}                     & \multicolumn{2}{c|}{25}                      & \multicolumn{2}{c|}{50}                      & \multicolumn{2}{c}{100} \\
    \midrule
    \textbf{Rank}                    & \textbf{c} & \multicolumn{1}{c|}{\textbf{p}} & \textbf{c} & \multicolumn{1}{c|}{\textbf{p}} & \textbf{c} & \multicolumn{1}{c|}{\textbf{p}} & \textbf{c} & \multicolumn{1}{c|}{\textbf{p}} & \textbf{c} & \multicolumn{1}{c|}{\textbf{p}} & \textbf{c}  & \textbf{p} & \textbf{c} & \multicolumn{1}{c|}{\textbf{p}} & \textbf{c} & \multicolumn{1}{c|}{\textbf{p}} & \textbf{c} & \multicolumn{1}{c|}{\textbf{p}} & \textbf{c} & \multicolumn{1}{c|}{\textbf{p}} & \textbf{c} & \multicolumn{1}{c|}{\textbf{p}} & \textbf{c} & \textbf{p} \\
    \textbf{0}                       & 4          & \multicolumn{1}{c|}{0.305}      & 4          & \multicolumn{1}{c|}{0.305}      & 4          & \multicolumn{1}{c|}{0.305}      & 4          & \multicolumn{1}{c|}{0.220}      & 4          & \multicolumn{1}{c|}{0.220}      & 5           & 0.270      & 3          & \multicolumn{1}{c|}{0.317}      & 3          & \multicolumn{1}{c|}{0.317}      & 3          & \multicolumn{1}{c|}{0.317}      & 10         & \multicolumn{1}{c|}{0.358}      & 11         & \multicolumn{1}{c|}{0.358}      & 27         & 0.420      \\
    \textbf{1}                       & 5          & \multicolumn{1}{c|}{0.256}      & 5          & \multicolumn{1}{c|}{0.256}      & 6          & \multicolumn{1}{c|}{0.282}      & 4          & \multicolumn{1}{c|}{0.179}      & 4          & \multicolumn{1}{c|}{0.179}      & 6           & 0.272      & 5          & \multicolumn{1}{c|}{0.334}      & 5          & \multicolumn{1}{c|}{0.334}      & 5          & \multicolumn{1}{c|}{0.334}      & 15         & \multicolumn{1}{c|}{0.345}      & 17         & \multicolumn{1}{c|}{0.346}      & 41         & 0.414      \\
    \textbf{2}                       & 5          & \multicolumn{1}{c|}{0.169}      & 5          & \multicolumn{1}{c|}{0.169}      & 5          & \multicolumn{1}{c|}{0.169}      & 1          & \multicolumn{1}{c|}{0.197}      & 1          & \multicolumn{1}{c|}{0.197}      & 2           & 0.331      & 1          & \multicolumn{1}{c|}{0.067}      & 1          & \multicolumn{1}{c|}{0.067}      & 1          & \multicolumn{1}{c|}{0.067}      & 6          & \multicolumn{1}{c|}{0.295}      & 8          & \multicolumn{1}{c|}{0.310}      & 30         & 0.422      \\
    \textbf{3}                       & 2          & \multicolumn{1}{c|}{0.094}      & 2          & \multicolumn{1}{c|}{0.094}      & 2          & \multicolumn{1}{c|}{0.094}      & 0          & \multicolumn{1}{c|}{N/A}        & 0          & \multicolumn{1}{c|}{N/A}        & 0           & N/A        & 1          & \multicolumn{1}{c|}{0.045}      & 1          & \multicolumn{1}{c|}{0.045}      & 1          & \multicolumn{1}{c|}{0.045}      & 2          & \multicolumn{1}{c|}{0.198}      & 2          & \multicolumn{1}{c|}{0.198}      & 4          & 0.337      \\
    \textbf{4}                       & 1          & \multicolumn{1}{c|}{0.003}      & 1          & \multicolumn{1}{c|}{0.003}      & 1          & \multicolumn{1}{c|}{0.003}      & 0          & \multicolumn{1}{c|}{N/A}        & 0          & \multicolumn{1}{c|}{N/A}        & 0           & N/A        & 0          & \multicolumn{1}{c|}{N/A}        & 0          & \multicolumn{1}{c|}{N/A}        & 0          & \multicolumn{1}{c|}{N/A}        & 1          & \multicolumn{1}{c|}{0.027}      & 1          & \multicolumn{1}{c|}{0.027}      & 1          & 0.027      \\
    \bottomrule
\end{tabular}
    \vskip -0.2cm
    \caption{
        The results of our quantitative BST evaluation.
        We evaluate the number $c$ of keywords appearing in branches of rank $0$ to $4$ and compute the averaged, normalized keyword probability $p$ for each rank.
        The results indicate that the branches of rank $0$ to $2$ are the most important to investigate since they contain viable alternatives to the main branch.
        Also, the probability only slightly decreases in the lower ranks.
    }
    \label{tab:quant-bst-eval}
    \vspace{-10pt}
\end{table*}

\subsection{Configuration and Text Generation}
To support the configuration task~\reflmtask{Conf}, the generAItor workspace allows loading pre-trained language transformers.
All generative language transformers from HuggingFace~\cite{Wolf2020TransformersStateArt} can be loaded and used.
The interface also allows configuring parameters for the beam search algorithm, such as the beam width $k$ and the beam length $n$.
Finally, the user can create prompts to be loaded into the workspace for text generation, implementing the text generation task~\reflmtask{Gen}.

\subsection{Beam Search Tree Visualization}
\begin{figure}[!tb]
    \centering
    \includegraphics[width=\linewidth]{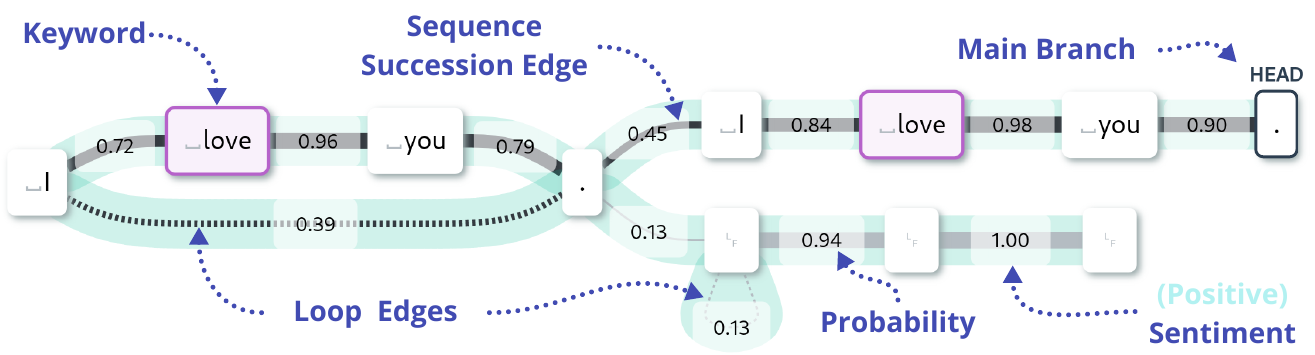}
    \caption{The beam search tree visualization.}
    \label{fig:loop}
    \vspace{-10pt}
\end{figure}

Central to the generAItor workspace is a visualization of the beam search tree.
As shown in~\cref{fig:loop}, we augment the tree with additional information, supporting the single-instance analysis task~\reflmtask{Single}.
The edges of the tree show alternative paths and encode the probability of the following nodes, which allows investigating surface form competition~\reflmchallenge{SFC}.
Semantic node highlights~\cite{ElAssady2022SemanticColorMapping} facilitate the identification of related keywords in the tree based on their high-dimensional token embeddings in the language model.
The edges are highlighted with the branch's sentiment to investigate the influence of negations~\reflmchallenge{Neg} or to analyze negative connotations through biased outputs~\reflmchallenge{Bias}.

\subsection{Comparative Tree Visualization}
Complementing the single-instance analysis, generAItor provides a second mode for comparing multiple tree instances.
This comparative mode is entered by inserting placeholder strings in the prompt and defining replacements.
Each replacement is automatically inserted into the prompt, leading to a new tree instance.
The instances are shown next to each other, facilitating comparison across multiple trees, enabling comparative analysis~\reflmtask{Multi}.
This allows the investigation of changes in the output, e.g., to probe different quantifiers~\reflmchallenge{Quant} or investigate prompt sensitivity~\reflmchallenge{Sens} by dynamically changing punctuation in the prompt.

\subsection{Highlighting and Abstraction}
To alleviate the complexity of the produced tree visualization, generAItor allows reducing the number of displayed nodes for close reading.
In particular, the user can specify a wordlist with interesting words for the analysis (or select one of the pre-defined wordlists). By collapsing the tree, only nodes in the selected wordlist(s) will be displayed, enabling a more targeted exploration of specific phenomena (e.g., stereotypical words). An example is shown in~\cref{fig:quantifiers}.

\section{Quantitative BST Evaluation}
\label{sec:eval-quant-bst}
In the following, we show the relevance of our tree-centered approach by evaluating how many relevant words are hidden in runner-up branches and would, therefore, be discarded in a usual text generation setting.
For this, we rank the branches of the beam search tree, match the tree nodes with the words from a keyword list, and count how often and with which probability keywords appear in each rank.

\begin{figure}[b!]
    \vspace{-10pt}
    \centering
    \includegraphics[width=\linewidth]{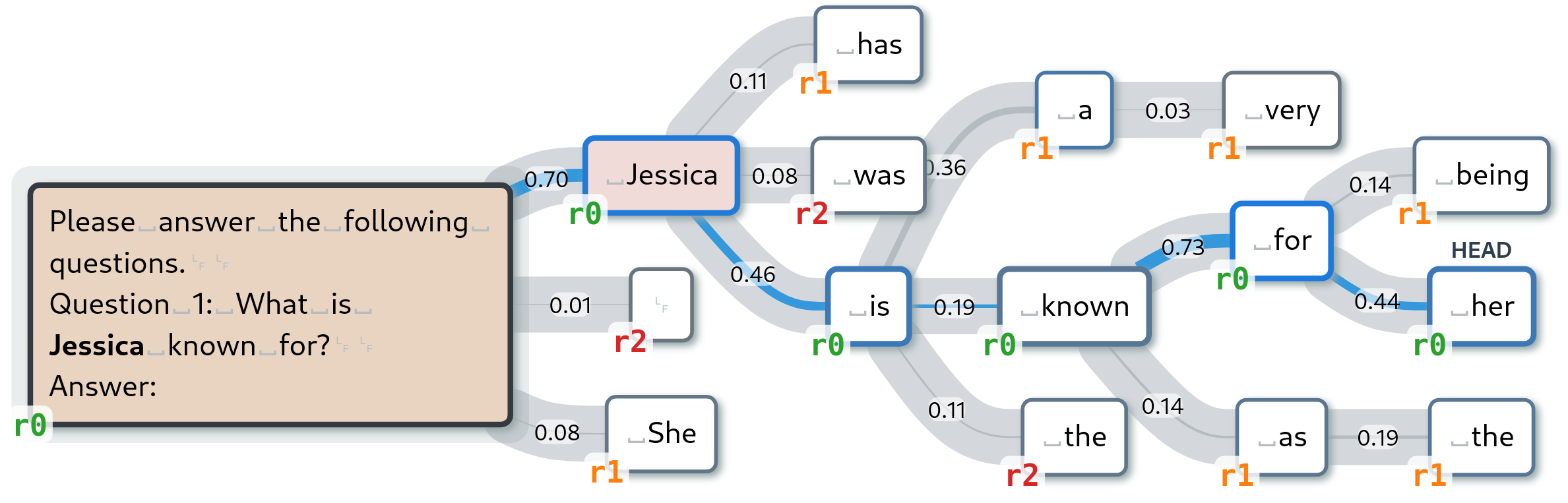}
    \caption{Example of applying~\cref{lst:ranking-algorithm} to a BST.}
    \label{fig:path-rankings}
\end{figure}

\paragraph{Ranking Beam Search Branches}
We require a ranking function on the branches of the beam search tree to determine their relevance.
Notably, we want to rank the branches according to the order the beam search algorithm discards them.
To this end, we propose ~\cref{lst:ranking-algorithm}.
\begin{lstlisting}[label={lst:ranking-algorithm}, caption={Ranking the branches of a BST.}, style=PythonCustomLst, xleftmargin=0cm, float=b, captionpos=b, aboveskip=-0.3cm, belowskip=-2.2em]
def get_best_leaf(n):
    return n.leafs.sort(
        key=lambda l: (l.max_beam_length, l.max_beam_prob),
        reverse=True)[0]

def rank(p):
    C = p.children.sort(
        key=lambda c: (get_best_leaf(c).max_beam_length,
                       get_best_leaf(c).max_beam_prob),
        reverse=True)
    for i, c in enumerate(C):
        c.rank = p.rank + i
        rank(c)

root.rank = 0
rank(root)
\end{lstlisting}
Intuitively, the algorithm assigns the lowest rank $0$ to the main branch of the beam search tree; then, at each branching point, the longest beam inherits its parent's rank, while the other branches receive a higher rank according to their order of being discarded.
\Cref{fig:path-rankings} shows an example ranking.

\paragraph{Evaluating Keyword Coverage}
We evaluate the keyword coverage for beam search trees produced with the models \textit{bloom-3b} and \textit{RedPajama-INCITE-Base-3B-v1} and different input prompts.
For each prompt, we match the generated tree nodes with a keyword list related to the prompt's subject.
E.g., we use a keyword list containing the names of all countries to match the generated output of the prompt \prompt{World economy is strongly dependent of some countries}.
The nodes of a branch are ranked according to~\cref{lst:ranking-algorithm}.
We then count the occurrences $c$ of keyword nodes in rank $0, 1, \ldots, k-1$, where $k$ is the beam width.
We also compute the normalized probability $p_{norm} = {p_{beam}}^{1 / d}$ of the keyword nodes, based on their beam probability $p_{beam}$ and depth $d$ in the tree.
This compensates for the exponential drop in probability as the beam length increases and allows us to compute an averaged probability $p$ of the keyword nodes in each rank.

\paragraph{Results}
The results of our experiment are depicted in~\cref{tab:quant-bst-eval}, showing that branches of rank $1$ contain the most keyword nodes, surpassing the number in the main branch with rank $0$.
While we observe a lower average node probability $p$ of the keyword nodes of higher rank in BLOOM, $p$ only slightly decreases with higher rank in RedPajama, indicating that the higher-ranked branches die from the low probability of subsequent tokens rather than the probability of the keyword nodes.

In summary, the results demonstrate the importance of a beam-search-tree-based approach.
Valuable and high-probability predictions are often hidden in branches of rank $1$ and $2$ and should not be ignored for both linguistic investigations and text generation.
Our results also show that examining BSTs with a beam width $k > 4$ may only rarely make sense since these branches tend to die early and hardly contain relevant keywords.

\section{Prompting Challenge Scenarios}
\label{sec:case-studies}
In the following, we present five example scenarios of how to use the generAItor workspace to examine the prompting challenges introduced in~\cref{sec:problem-characterization-and-methodology}.

\subsection{Scenario: Prompt Sensitivity}
\label{subsec:prompt-sensitivity}
\custombox{SummaryBoxColor}{Scenario: Influence of Punctuation}{%
    \contentrow{Model}{\small RedPajama-INCITE-Instruct-3B-v1}
    \contentrow{Prompt}{\small\texttt{Answer the following questions.\newline
    Q: What is the current GDP of India?\newline
    A:<PH>}}
    \contentrow{<PH>}{\small\textit{\{\}}\texttt{,\hquad\textvisiblespace,\hquad\textvisiblespace\textvisiblespace}}
    \contentrow{Challenge}{Prompt Sensitivity~\reflmchallenge{Sens}}
    \contentrow{Task}{Multi-Instance~\reflmtask{Multi}}
}

\begin{figure}[t]
    \centering
    \includegraphics[width=\linewidth]{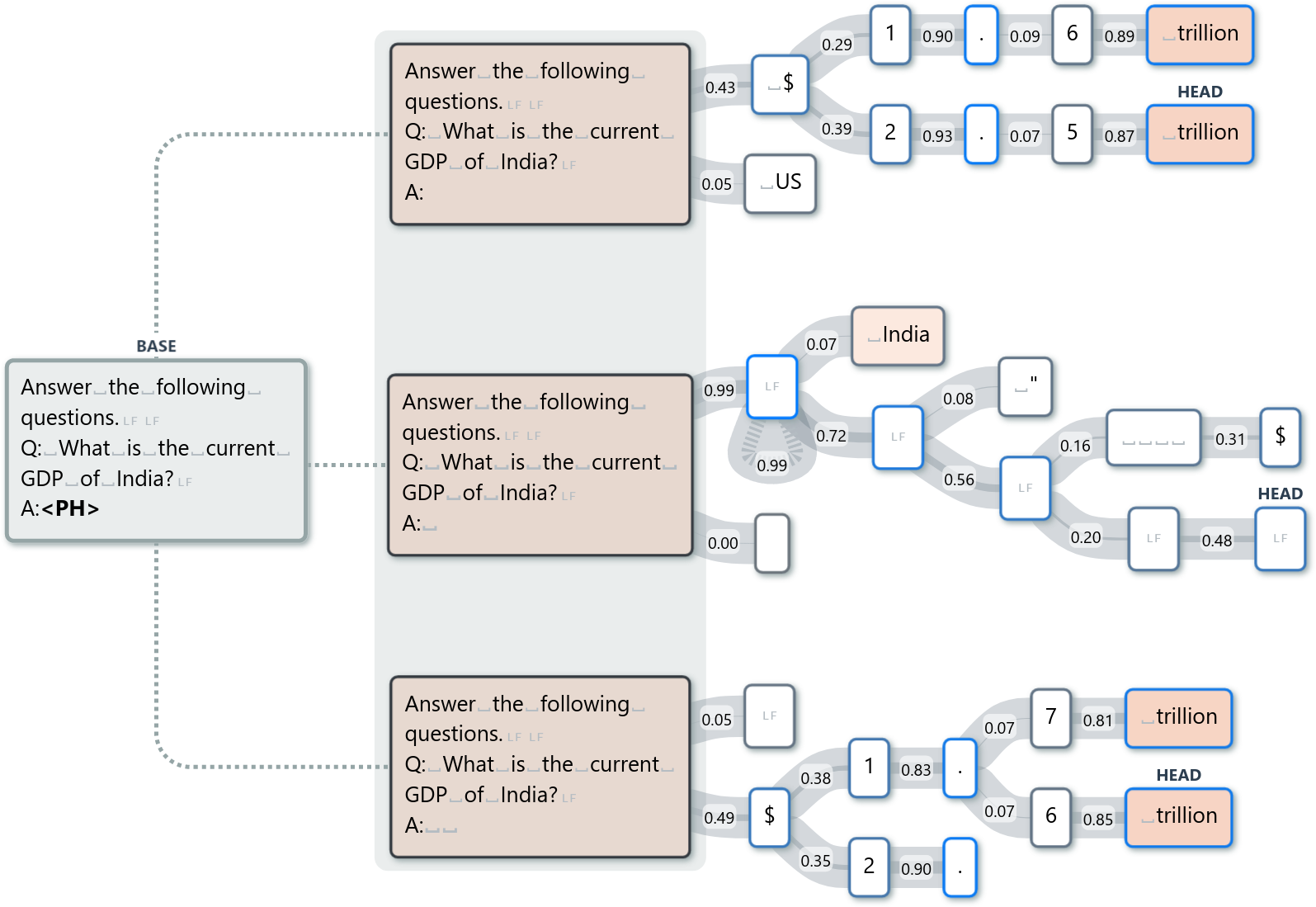}
    \caption{A comparative BST, showing how strongly punctuation in the input prompt influences the outputs.}
    \label{fig:india-gdp}
    \vspace{-5pt}
\end{figure}

\noindent
In this scenario, we show how our workspace can be used to analyze prompt sensitivity to minor adaptations.
In particular, we show the sensitivity of the RedPajama Instruct model to white spaces added to the input prompt.
We use the prompt \prompt{Answer the following questions. Q: What is the current GDP of India? A:<PH>} whereby the \prompt{<PH>} stands for 0--2 concatenated white spaces (i.e., the prompt starts with either \prompt{\hquad},\hquad\prompt{\textvisiblespace}, or\hquad\prompt{\textvisiblespace\textvisiblespace}).
As shown in~\cref{fig:india-gdp}, the model generates three unique BST trees, each containing a unique text output.
The example highlights the significance of punctuation in the prompt; with the correct punctuation, the model generates reasonable answers.
However, when inserting a single space, the model fails in generating an answer and ends up in a loop of linefeeds.
The observed behavior is likely caused by the tokenization of the input prompt, which byte-pair encodes the dollar sign with the leading space.
Then, the model is trained to expect the combined \prompt{\textvisiblespace\$} preceding the answer.
Besides prompt sensitivity, this example also highlights the importance of investigating probabilities of alternative branches, as both branches exiting the root node of the tree at the top have similar probabilities, indicating likely hallucinations.

\subsection{Scenario: Surface Form Competition}

\custombox{SummaryBoxColor}{Scenario: Surface Form Competition}{%
    \contentrow{Models}{\small gpt2, RedPajama-INCITE-Base-3B-v1}
    \contentrow{Prompt}{\small\texttt{A human wants to submerge himself in water, what should he use?\newline
    Possible answers are: "Coffee cup", "Whirlpool bath", "Cup", "Puddle"\newline
    Answer: "}}
    \contentrow{Challenge}{Surface Form Competition~\reflmchallenge{SFC}}
    \contentrow{Task}{Single-Instance~\reflmtask{Single}}
}

\begin{figure}[t]
    \centering
    \includegraphics[width=\linewidth]{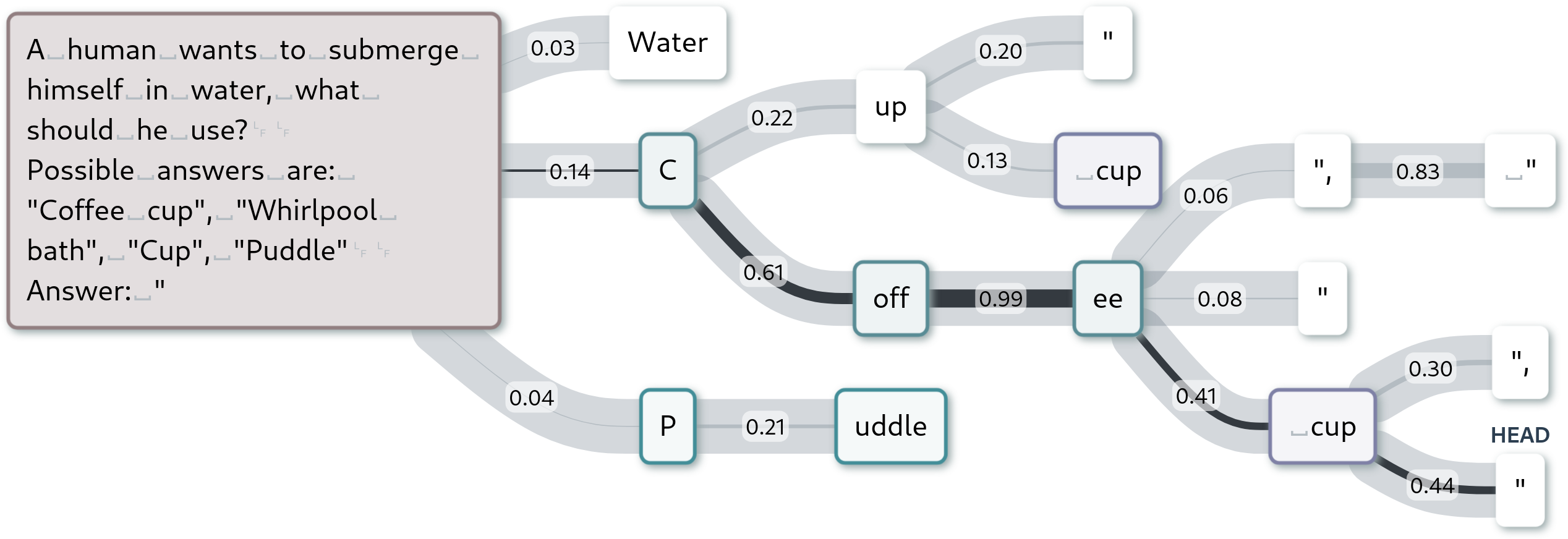}
    \caption{The BST for the example from~\citet{Holtzman2021SurfaceFormCompetition}, showing how surface form competition affects the output probabilities.}
    \label{fig:sfc}
    \vspace{-10pt}
\end{figure}

\noindent
In this scenario, we show how our workspace is used to analyze surface form competition using the prompt \prompt{A human wants to submerge himself in water, what should he use? Possible answers are: "Coffee cup", "Whirlpool bath", "Cup", "Puddle" Answer: "} from~\citet{Holtzman2021SurfaceFormCompetition}.
Our tree confirms that the most likely result is not the correct answer \prompt{Whirlpool bath}, but the hallucinations \prompt{Coffee cup} for GPT-2 and \prompt{Cup} for RedPajama Base.

It should be noted that we also tried other examples from the paper, e.g., the prompt \prompt{What is the most populous nation in North America? Valid answers: "U.S. of A.", "Canada" Answer: "}.
However, we were not able to reproduce the results from the paper, as both GPT-2 and RedPajama Base rated \prompt{U.S. of A.} more likely than \prompt{Canada}.

\subsection{Scenario: Negation}

\custombox{SummaryBoxColor}{Scenario: Negation}{%
    \contentrow{Model}{\small RedPajama-INCITE-Instruct-3B-v1}
    \contentrow{Prompt}{
        \small\texttt{Answer my questions. Do not use the word `strawberries`.\newline
        Q: Which type of red berries grows on small, green bushes?\newline
        A:}
        \newline \vskip 0.25em
        \texttt{Answer my questions. Do not use the word `raspberries`.\newline
        Q: Which type of red berries grows on small, green bushes?\newline
        A:}
    }
    \contentrow{Challenge}{Negation~\reflmchallenge{Neg}}
    \contentrow{Task}{Single-Instance~\reflmtask{Single}}
}

\begin{figure}
    \centering
    \includegraphics[width=\linewidth]{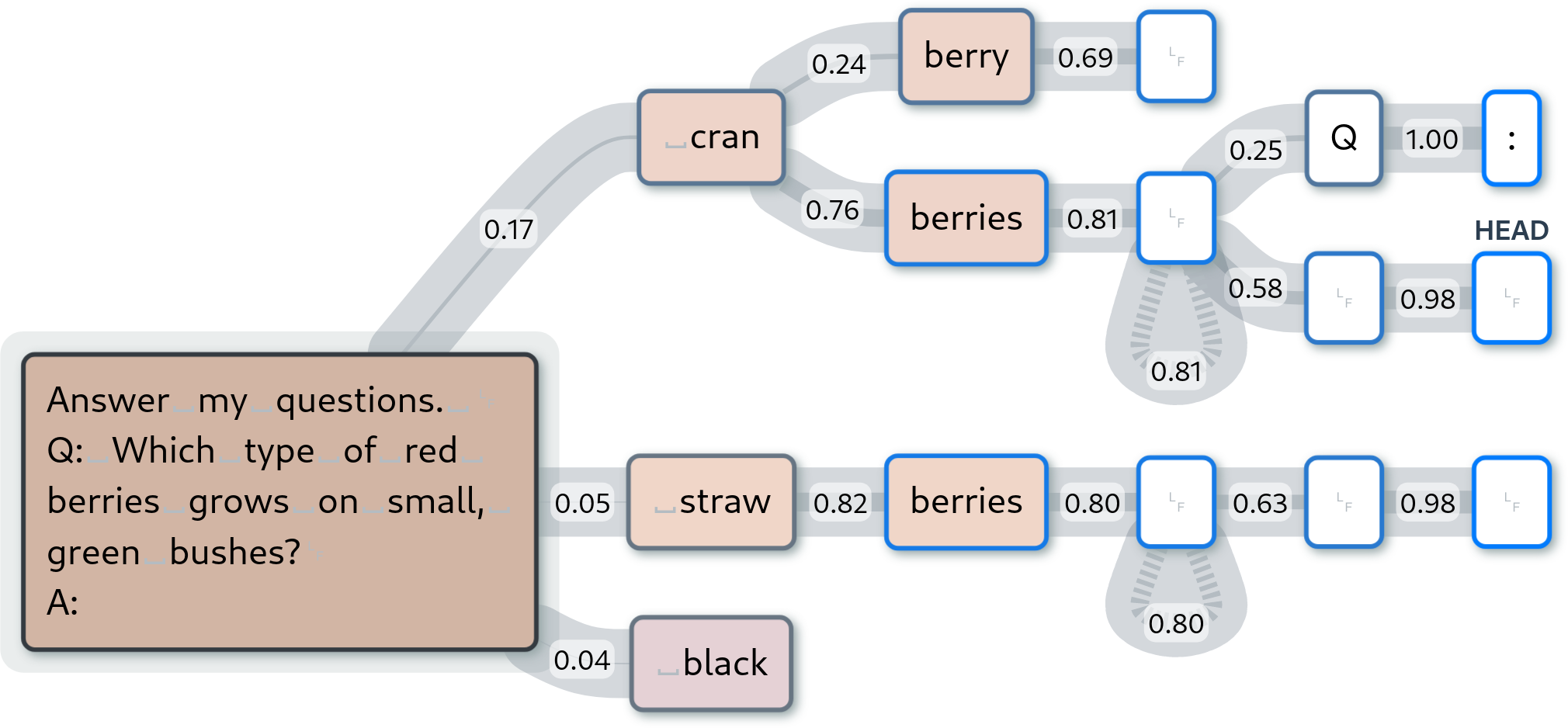}
    \caption{The baseline for the negation analysis: the token \prompt{raspberries} is not among the top-$3$ predictions.}
    \label{fig:raspberries}
    \vspace{-15pt}
\end{figure}

\begin{figure}[b!]
    \vspace{-10pt}
    \centering
    \includegraphics[width=\linewidth]{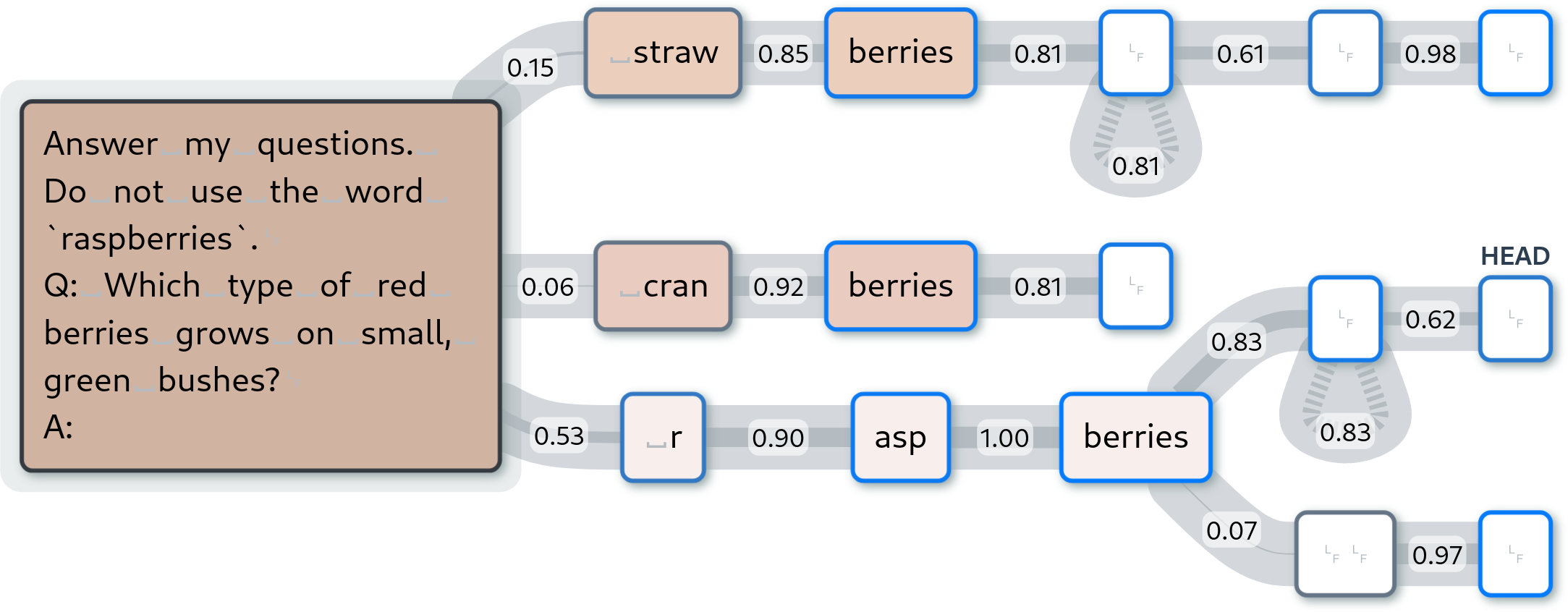}
    \caption{A BST showing how the negative imperative \prompt{do not use} boost the probability of the unwanted token.}
    \label{fig:raspberries-not}
\end{figure}

\noindent
In this scenario, we investigate how RedPajama's Instruct model captures the semantic constraints of the negation \prompt{not}.
First, we aim to explore the most likely prediction for the prompt \prompt{Answer my questions. Q: Which type of red berries grows on small, green bushes? A:}.
The model predicts multiple berry types including cranberries and strawberries, shown in~\cref{fig:raspberries}.
Since these predictions do not include the word \prompt{raspberries}, we use it to verify whether the model can interpret the meaning of \prompt{not}.
Thus, we additionally create a prompt \prompt{Answer my questions. Do not use the word `raspberries`.
Q: Which type of red berries grows on small, green bushes?
A:}.
If the model can interpret the meaning of the negation, the predictions should not include the word \prompt{raspberries}.
However, the model ranks this word as the most likely one, see~\cref{fig:raspberries-not}, from which we conclude that the model does not capture the semantic constraints of the negation.

\subsection{Scenario: Quantifiers}

\custombox{SummaryBoxColor}{Scenario: Quantifiers}{%
    \contentrow{Model}{\small gpt2, bloom-3b}
    \contentrow{Prompt}{\small\texttt{<PH> women like to}}
    \contentrow{<PH>}{\small\texttt{All},\hquad\texttt{Some},\hquad\texttt{A few}}
    \contentrow{Challenge}{Quantifiers~\reflmchallenge{Quant}}
    \contentrow{Task}{Multi-Instance Analysis~\reflmtask{Multi}}
}
\begin{figure*}
    \centering
    \includegraphics[width=\textwidth]{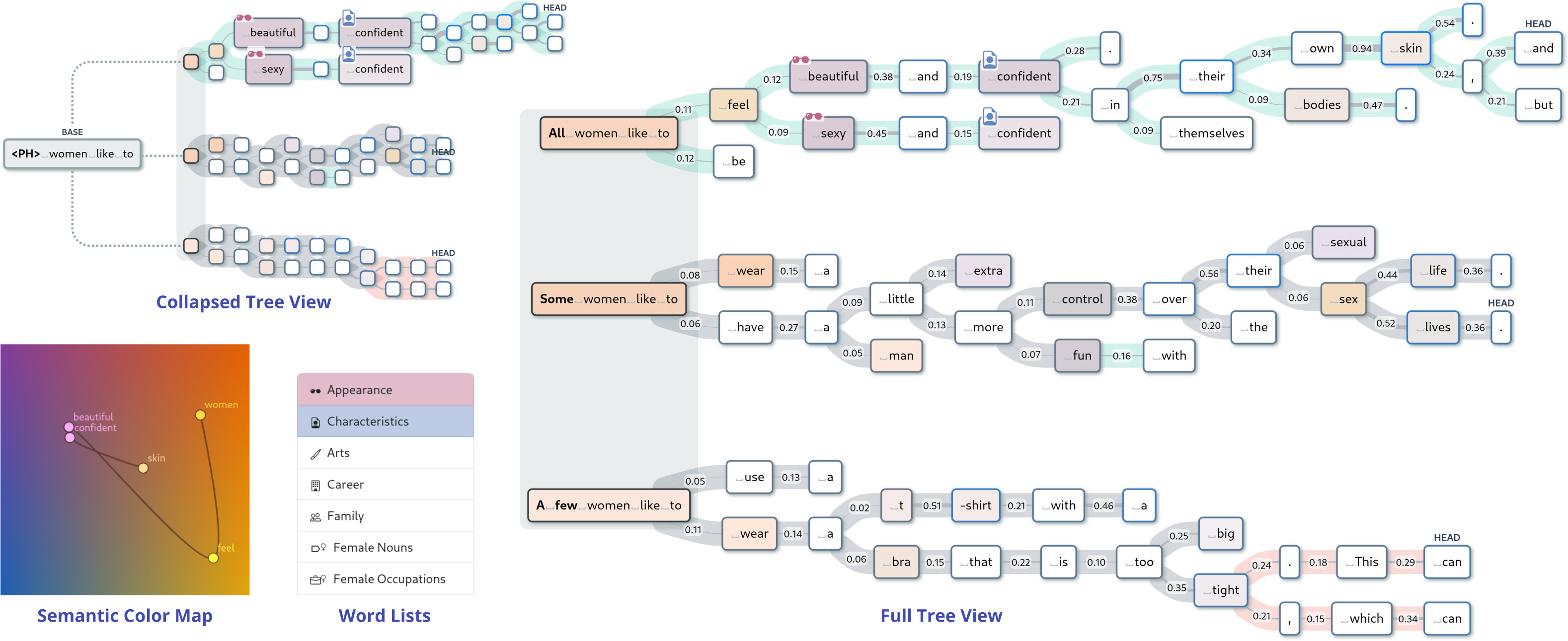}
    \caption{The BSTs for the prompt \prompt{<PH> women like to} with different quantifiers used in the place of the \prompt{<PH>} token. The user can select wordlists for exploration; the tree is collapsed showing only interesting nodes for the analysis.}
    \label{fig:quantifiers}
    \vspace{-10pt}
\end{figure*}
\noindent
In the following, we explore how language models encode quantifiers such as \prompt{all}, \prompt{some}, and \prompt{a few}.
\citet{Gupta2023ProbingQuantifierComprehension} shows that larger generative models are able to learn the semantic constraints of these function words better than smaller models or masked language models~\cite{Kalouli2022NegationCoordinationQuantifiersa}.
We explore the ability of GPT-2 and BLOOM to capture these properties using the prompt  \prompt{<PH> women like to} whereby the  \prompt{<PH>} stands for the placeholder for words \prompt{all}, \prompt{some}, and \prompt{a few}.
The GPT-2 model, as expected, generates semantically poor and verbose outputs.
The prompts that include the word \prompt{all} and \prompt{a few} produce the same top prediction, i.e., the model generates a sequence \prompt{<PH> women like to think that they are the only ones who have the power to change the world}.
As shown in~\cref{fig:quantifiers}, the predictions of BLOOM differ from GPT-2.
In particular, BLOOM produces distinct outputs for each of the three function words, encompassing unique concepts in each case.
This confirms the findings by~\citet{Gupta2023ProbingQuantifierComprehension} that larger models generate outputs that address the quantifiers better.
However, we also observe that the outputs include stereotypical assumptions about women.
Especially for the quantifier \prompt{all}, the predictions overemphasize the relevance of aesthetics to the female gender (see \prompt{All women like to feel beautiful and confident in their own skin.} in~\cref{fig:quantifiers}).
In the following, we describe in more detail how our approach helps in investigating biases encoded in the model's parameters.

\subsection{Scenario: Bias}
\label{subsec:case-study-comparative-analysis}

\custombox{SummaryBoxColor}{Scenario: Bias}{%
    \contentrow{Model}{\small bloom-3b}
    \contentrow{Prompt}{\small\texttt{<PH> women like to}}
    \contentrow{<PH>}{\small\texttt{All}, \texttt{Some}, \texttt{A few}}
    \contentrow{Challenge}{Bias~\reflmchallenge{Bias}}
    \contentrow{Task}{Multi-Instance~\reflmtask{Multi}}
}

\noindent
As shown in~\cref{fig:quantifiers}, the predictions for the prompt  \prompt{<PH> women like to} with words \prompt{all}, \prompt{some}, and \prompt{a few} in the place of the placeholder \prompt{<PH>} produce stereotypical predictions.
Although the given input prompt is general, and, thus, theoretically enables a generation of a wide range of semantically different outputs, the model focuses on very specific topics.
In particular, in addition to the aesthetic aspects associated with the prompt \prompt{All women like to}, the other prompts produce predictions that contain properties related to female body characteristics (see~\cref{fig:quantifiers}).

\section{Discussion \& Take-Home Messages}
In the following, we discuss our work and derive the most important take-home messages.

\paragraph{Visual, Qualitative Analysis}
Our case studies highlight the importance of inspecting the prompt output differences visually.
Visualizations are often used to gain detailed insights into specificities that might become opaque when applying solely quantitative evaluation approaches (e.g., accuracy scores).
Visualizations can be especially useful to test assumptions since such tests are cheap to execute. The gained insights can then be used to define hypotheses that are evaluated quantitatively.

\paragraph{Comparative Analysis}
Comparative analysis, i.e., the possibility to compare the outputs for multiple prompts simultaneously is crucial to detect model limitations.
Often, only the relative difference to another prompt can reveal the cues to which the model pays attention, to which aspect it is sensitive, and which linguistic properties are not considered for the prediction making.

\paragraph{Simplicity}
Since language is inherently interpretable \cite{Sevastjanova2022BewareRationalizationTrap}, individuals are led to engage in a process of rationalizing language model outputs.
Interestingly, studies have shown that users tend to place trust in the explanations provided by language models, even in cases where those explanations are proven to be incorrect~\cite{Lai2019HumanPredictionsExplanations}.
To address this issue, our approach exposes the BST, thereby offering an inherent explanation of the model outputs.
The fundamental principle underlying our approach lies in the simplicity of both the beam search algorithm and the underlying data, such as token probabilities.
This simplicity helps prevent the occurrence of misleading rationalizations concerning the generated predictions.

\paragraph{Flexibility \& Abstraction}
The analysis of language model outputs using the BST enables the expansion of sequences to variable lengths, which distinguishes it from template-based analysis. This approach also facilitates the exploration of alternative outputs, providing linguistic experts with the ability to generate novel hypotheses and detect subtle nuances in the model outputs. For instance, it allows for identifying biases present in longer sequences rather than being limited to static n-grams.
Overall, the BST-based analysis empowers users to gain deeper insights into the model's behavior and uncover more intricate patterns within its outputs.
To ensure scalability, it is crucial to employ effective abstraction techniques (such as tree collapse or keyword highlights) that prevent users from getting overwhelmed by the vast exploration space.

\section{Related Work}
\label{sec:related-work}
In the following, we present related work on language modeling, language model explainability, and beam-search-tree-based visualizations.

\subsection{Language Modeling}
LMs are probability distributions over word sequences and a core component of natural language processing (NLP) systems~\cite{Bengio2000NeuralProbabilisticLanguage}.
With the emergence of the transformer architecture~\cite{Vaswani2017AttentionIsAll}, LM research shifted away from using recurrent neural networks~\cite{Rumelhart1986LearningRepresentationsBack} due to the inherent parallelism of transformers that decreases training times and provides superior performance in capturing long-term dependencies as a result of utilizing attention mechanisms~\cite{Bahdanau2014NeuralMachineTranslation}.

Among LMs, two main types can be distinguished: masked models (e.g., BERT~\cite{Devlin2019BertPreTraining}) and generative models (e.g., GPT-2~\cite{Radford2019LanguageModelsAre}).
In this paper, we focus on text generation, which is best tackled by using autoregressive generative models that are trained to predict the next token following an input sequence~\cite{Li2021PretrainedLanguageModel}.
For our case studies, we use GPT-2, BLOOM~\cite{Scao2023Bloom176bParameter} and RedPajama~\cite{Computer2023RedpajamaOpenSource}, but note that the models can be exchanged by any other causal transformer LM.

\subsection{Language Model Explainability}

With the rise of large language models, the explainability of their inner workings and the interpretability of their outputs expanded the field of explainable AI.
Matching the four categories as proposed by \citet{Danilevsky2020SurveyStateExplainable}, approaches usually use explainability techniques in conjunction with a set of operations to enable explainability, and visualization techniques to convey the operations to the user.
Examples are visualizing saliency to explain feature importance for local post-hoc~\cite{Mullenbach2018ExplainablePredictionMedical} or training a surrogate model to allow for taxonomy induction, providing global explanations~\cite{Liu2018InterpretationNetworkEmbedding}.

As identified by \citet{Yuan2020SurveyVisualAnalytics}, explanations are needed before, during, and after model building, and it is crucial to identify ways to intuitively convey model outputs to the user and allow for an exploration of model outputs.
In the context of visual analytics approaches for the explainability of deep neural networks,
\citet{Rosa2023StateArtVisual} survey common visualization techniques used in visual analytics systems for explainability and identify a lack of tree-based visualization techniques.
Our proposed method is based on a representation of the beam search tree and complements it with a set of interactions for example-driven, instance-based investigation of NLP challenges, offering both self-explaining and post-hoc local explanations.

\subsection{Beam-Search-Tree-Based Visualizations}

Beam search is an essential part of the decoding process in LMs. Visualizing and using the created beam search tree is, therefore, a possibility to investigate predictions and allow user interaction with the tree.
\citet{Lee2017InteractiveVisualizationManipulation} use a basic beam search tree visualization for the task of neural machine translation. Their tool visualizes the beam search decoder with probabilities and allows basic tree manipulation.
Also, for machine translation, Seq2Seq-Vis was proposed by \citet{Strobelt2018Seq2seqVisVisual}, which focuses on helping the user debug and find errors in the translation result. The user can investigate all steps of the translation pipeline to help improve the translation result for single instances.
For larger document collections, \citet{Munz2022VisualizationBasedImprovement} propose a visual analytics system to help identify and correct single instances and propagate corrections for larger document collections.
They also visualize the beam search tree and allow basic interactions on the node level to correct translations.
\citet{Strobelt2022GenniHumanAi} introduce GenNI, a system for collaborative text generation by applying user-defined constraints to the beam search tree, guiding the produced outputs.

\section{Conclusion}
We present a beam-search-centered approach to explainability for (and comparison of) generative language models by putting the beam search tree in the center of the generAItor visual analytics technique.
For this technique, we leverage the beam search tree to explain the model's decision process and compare model outputs.
Using our approach, we find that state-of-the-art LMs handle quantifiers well, while at the same time producing strongly biased output.
Our investigation of negations highlights how it is ignored by the tested models, as including a negative imperative in the prompt boosts the probability of the unwanted output instead of decreasing it.

Overall, we tackle five prototypical prompting challenges to highlight how the visual investigation of probabilities and alternative branches aids in verifying and generating hypotheses for LM developers and linguistic researchers alike.

\section*{Limitations}
\paragraph{Investigation of Proprietary Models}
Since our approach requires full access to the probability distribution output by the model, it can only be applied to open-source models.
However, similar approaches could be included in commercial tools for language generation, as prompt engineering is gaining relevance~\cite{ZamfirescuPereira2023WhyJohnnyCant}.
Gaining insights into the generated outputs has the potential to greatly enhance human control.

\paragraph{Comparison Across Language Models}
While our approach allows loading different, transformer-based models into the workspace, the comparison of outputs is at present only supported between trees produced by the model that is currently loaded.
This limitation should be supported by future implementations.

\paragraph{Focus on the English Language}
Due to the prevalence of English training data, most models are known to provide the best performance with English text.
We, therefore, focus on English text for the examples and evaluations presented in this paper.
Since the linguistic phenomena we examine can strongly differ between languages, further languages should be investigated in future work.

\paragraph{Extension to further Prompt Challenges}
The identified and addressed prototypical challenges represent current areas of active research.
Nevertheless, it is likely that there are further interesting linguistic, socio-linguistic, or data- and model-specific prompting challenges that can be investigated using the generAItor workspace.

\paragraph{Focus on Text Generation}
Other tasks, such as machine translation or text summarization were not investigated.
While our approach technically supports these tasks, additional visualizations and interaction patterns may have to be implemented to optimally support the user and should be part of future research.

\paragraph{Explainability Instead of Problem Solving}
While some of our insights indicate model defects and imply ways to resolve them (e.g., preventing tokenization issues, see~\ref{subsec:prompt-sensitivity}), this is not the primary focus of our approach.
To find tangible ways to refine a model, other tools to investigate training data or the deep learning architecture of the model are needed.

\clearpage
\bibliographystyle{acl_natbib}

% FOR ARXIV SUBMISSION:
% "We do not run BibTeX in the auto-TeXing procedure. If you use it,
%  include in your submission the .bbl file that BibTeX produces on
%  your own machine."
\bibliography{main.bbl}

\end{document}